\documentclass[%
reprint,
 amsmath,amssymb,
 aps,
]{revtex4-2}

\usepackage{graphicx}
\usepackage{dcolumn}
\usepackage{bm}
\usepackage{physics}
\usepackage{float}
\usepackage{amsmath}
\usepackage{xcolor, soul}
\usepackage{multirow}

\begin{document}

\preprint{APS/123-QED}

\title{A Hybrid Quantum-Classical Neural Network Architecture for Binary Classification}

\author{Davis Arthur}
 \affiliation{University of Florida, Florida, USA}
 \email{davisarthur@ufl.edu}


\author{Prasanna Date}
\affiliation{
 Oak Ridge National Laboratory, Tennessee, USA
}
\email{datepa@ornl.gov}


\date{\today}

\begin{abstract}
Deep learning is one of the most successful and far-reaching strategies used in machine learning today. However, the scale and utility of neural networks is still greatly limited by the current hardware used to train them. These concerns have become increasingly pressing as conventional computers quickly approach physical limitations that will slow performance improvements in years to come. For these reasons, scientists have begun to explore alternative computing platforms, like quantum computers, for training neural networks. In recent years, variational quantum circuits have emerged as one of the most successful approaches to quantum deep learning on noisy intermediate scale quantum devices. We propose a hybrid quantum-classical neural network architecture where each neuron is a variational quantum circuit. We empirically analyze the performance of this hybrid neural network on a series of binary classification data sets using a simulated universal quantum computer and a state of the art universal quantum computer. On simulated hardware, we observe that the hybrid neural network achieves roughly 10\% higher classification accuracy and 20\% better minimization of cost than an individual variational quantum circuit. On quantum hardware, we observe that each model only performs well when the qubit and gate count is sufficiently small.

\end{abstract}

\maketitle


\section{Introduction}
\label{sec:intro}

Machine learning has revolutionized the modern world. Today, machine learning models are leveraged for nearly every imaginable task ranging from medical diagnoses \cite{Myszczynska2020} to fraud detection \cite{Awoyemi2017} to marketing \cite{Sterne2017}. The emergence of machine learning across many disciplines is due in large part to the recent accessiblity of relatively powerful computers. In accordance with Moore's Law, computer hardware has improved exponentially in scale and speed over the past 60 years. Unfortunately, despite all of the recent success, modern hardware still greatly restricts the practicality of certain machine learning models. Machine learning, deep learning in particular, can be very computationally expensive, sometimes requiring hours, days, or even months of training time on today's computers \cite{Thompson2020}. Moreover, conventional computers are beginning to approach physical limitations that will slow their improvements in years to come \cite{Peper2017}. For these reasons, many are beginning to research alternative computing platforms for training machine learning models. Among these platforms, quantum computers have emerged as a particularly interesting candidate.

The appeal of a quantum computer is largely due to the properties of quantum entanglement and quantum superposition which cannot be efficiently simulated on a classical computer. These properties can be extremely useful as illustrated by Shor's prime factorization algorithm \cite{Shor1999} and Grover's search algorithm \cite{Grover1996} which offer an exponential and polynomial speed up respectively over their best existing classical counterparts. These two algorithms give a sense of what large, high fidelity quantum computers may offer to the field of computer science in years to come. Today we are still in the era of noisy intermediate scale quantum (NISQ) computers, but already a number of quantum machine learning algorithms have been proposed \cite{Biamonte2017, Ciliberto2018}. We have previously studied quantum approaches to linear regression \cite{Date2020LinReg},  support vector machines \cite{Date2021QUBOFormulations}, and balanced $k$-means clustering \cite{Arthur2020balanced}. Additionally, we proposed a quantum learning model that can be used for binary classification on universal quantum computers \cite{Date2020QuantumDiscriminator}. In this paper, we propose a hybrid quantum-classical neural network architecture and empirically analyze its performance on several binary classification data sets. Our study is the first to test this hybrid neural network architecture using simulated or quantum hardware to the best of our knowledge.

Neural networks have proven to be successful for learning on conventional computers, but they have some serious limitations. They are prone to overfitting \cite{Hawkins2004}, and training even a small neural network is an NP-complete problem \cite{Blum1992}. These limitations have inspired many to propose quantum approaches to deep learning \cite{Schuld2015, Wan2017, Killoran2019, Beer2020, Zoufal2019, Kamruzzaman2019, Garg2020}. Unfortunately, many of these proposals cannot be implemented on modern hardware, and those that can, do not have a one-to-one correspondence with conventional artificial neural networks. For this reason, there is an increased interest in deep learning approaches that perform well on near term quantum computers. To this end, variational quantum circuits (VQCs) have proven to be a promising quantum analogue to artificial neurons \cite{Cerezo2021, Abbas2021, Benedetti2019, Broughton2020, Sim2019, Hubregtsen2021, Chen2020}. Variational quantum circuits can be trained using classical optimization techniques, and they are believed to have some expressibility advantages over conventional neural network architectures \cite{Cerezo2021}. Liu et al. have implemented their own hybrid quantum-classical neural network using variational quantum circuits, and their study provides detailed insight into the training dynamics of models similar to ours \cite{liu2021representation}.

\section{Quantum Neural Networks}

A variational quantum circuit is comprised of three key components. First, a feature map $F$ maps a real valued classical data point $\bm{x}$ into a $d$ qubit quantum state $\ket{\psi}$:
\begin{align}
    \ket{\psi(\bm{x})} = F(\bm{x}) \ket{0}^{\otimes d}
\end{align}
Next, an ansatz $A$ manipulates the prepared quantum state through a series of entanglements and rotation gates. The angles of the ansatz's rotations are parameterized by a vector $\bm{\theta}$.
\begin{align}
    \ket{\phi(\bm{x}, \bm{\theta})} = A(\bm{\theta}) \ket{\psi(\bm{x})} 
\end{align}
Finally, an observable $O$ is measured, and the eigenvalue corresponding to the resultant quantum state is recorded. In most machine learning applications, a variational quantum circuit is run many times using a particular input $\bm{x}$ and parameter vector $\bm{\theta}$ so that the circuit's expectation value, denoted by $f$, can be approximated. 
\begin{align}
    f(\bm{x}, \bm{\theta}) = \bra{\phi (\bm{x}, \bm{\theta})} O \ket{\phi (\bm{x}, \bm{\theta})}
\end{align}
When a variational quantum circuit is used for machine learning, this approximated expectation value is typically treated as the output of the model.
\begin{figure}[H]
    \centering
    \includegraphics[width=8 cm]{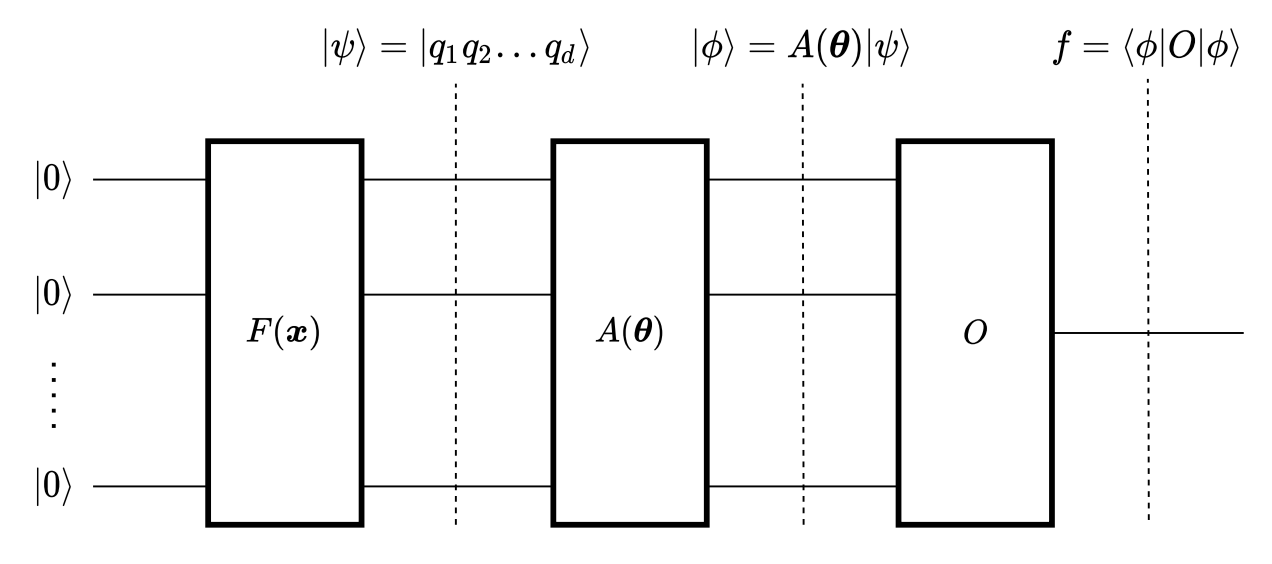}
    \caption{Schematic of a variational quantum circuit.}
    \label{fig:quantum_perceptron}
\end{figure}

The feature map of a variational quantum circuit is known to play some role in the expressiveness of the model \cite{Schuld2021FeatureMap}. In general, data should be encoded in such a way that the value of a feature can be extracted from the prepared quantum state through some combination of qubit rotations and measurements. This ensures each possible input $\bm{x}$ has a unique qubit encoding before being passed to the ansatz. On modern hardware, it is also important to use a feature map with limited depth, since each additional gate introduces noise to the quantum state. We satisfy both requirements by scaling each feature $x_i$ to fit within the interval $[0, \pi]$ and then encoding its value into the relative amplitude of a corresponding qubit:
\begin{align}
    \ket{q_i} = \cos \left( \frac{x_i}{2} \right) \ket{0} + \sin \left( \frac{x_i}{2} \right)  \ket{1}
\end{align}
It is worth noting that more sophisticated feature maps exist \cite{Goto2020, lloyd2020featureencoding, Yano2020}. However, for the data sets analyzed in this manuscript, this straightforward feature map achieves high accuracy while avoiding many of the complications introduced by more complex methods.

The Qiskit circuit library contains several ansatzes consisting of two qubit entanglements and parameterized single qubit rotations. We chose the \textit{RealAmplitudes} ansatz (with one repetition and full entanglement) for each variational quantum circuit studied in Section \ref{sec:results}. This is the default ansatz used by Qiskit's variational quantum circuit implementation (\textit{TwoLayerQNN}). It has also been used in a variational quantum circuit with proven advantages over traditional feedforward neural networks in terms of both capacity and trainability \cite{Abbas2021}. We also chose the default observable used by Qiskit's variational quantum circuit implementation. Mathematically, this observable can be described as the tensor product of $d$ Pauli-Z matrices ($\sigma_z$), where $d$ is the number of qubits in the quantum state:
\begin{align}
    O = \sigma_z^{\otimes d}
\end{align}
This observable has the interesting property that if the measured quantum state has odd parity, the recorded eigenvalue is -1, and if the measured quantum state has even parity, the recorded eigenvalue is 1. This means that the expectation value of the circuit will always be within the interval $[-1, 1]$. 

A number of studies have used a variational quantum circuit for binary classification \cite{Schuld2020BinaryClassification, Chen2021, Chen2020HybridClassifier, Farhi2018classification, Mitarai2018}. This can be done by relating the expectation value of the circuit to the probability that a point belongs to a given class. Consider a binary classification problem in which each data point $\bm{x}$ is labeled $y=1$ or $y=-1$. We use the following equation to relate the expectation value of the parity observable to the probability a point $\bm{x}$ is labelled $y$:
\begin{align}
    P(y|\bm{x}) = \frac{y f(\bm{x}, \bm{\theta}) + 1}{2}
    \label{eq:prob_perceptron}
\end{align}
Training the variational quantum circuit classifier amounts to determining a parameter vector $\bm{\theta}$ that minimizes the negative log-likelihood of the probability distribution over the training data set. The exact cost function used by our binary classifier is given by the equation below:
\begin{align}
    \text{Cost} = - \frac{1}{N} \sum_{i=1}^N \log{(  P(y_i|\bm{x}_i) )}
    \label{eq:cost}
\end{align}
where $N$ is the number of points in the training data set, $\bm{x}_i$ is the $i$th data point in the training set, and $y_i$ is the label of the $i$th point. 

Cost can be minimized using a classical optimizer such as gradient descent. When computing the gradient of the cost function, the derivative of the expectation value of the variational quantum circuit with respect to each parameter of the ansatz is computed using parameter shift rule \cite{Crooks2019}:
\begin{align}
    \dv{f}{\theta_i} = \frac{f(\theta_i + s) - f(\theta_i - s)}{2}
\end{align}
where $s$ is a macroscopic shift determined by the eigenvalues of the gate parameterized by $\theta_i$. For all of the rotation and phase gates available in the Qiskit library, $s = \pi / 2$. 

In many ways, the aforementioned variational quantum circuit classifier resembles a logistic unit used in a conventional neural network. The circuit has an input vector $\bm{x}$ and a set of classically optimizable parameters $\bm{\theta}$. Additionally, the output (expected value) of the variational quantum circuit is continuously differentiable and bound to a small range of real values. These similarities motivated us to construct a small hybrid quantum-classical feedforward neural network using variational quantum circuits as individual neurons. To achieve reasonable training times on modern quantum hardware, we restricted the neural network architecture to contain only a single hidden layer and a single output unit.
\begin{figure}[H]
    \centering
    \includegraphics[width=8.5cm]{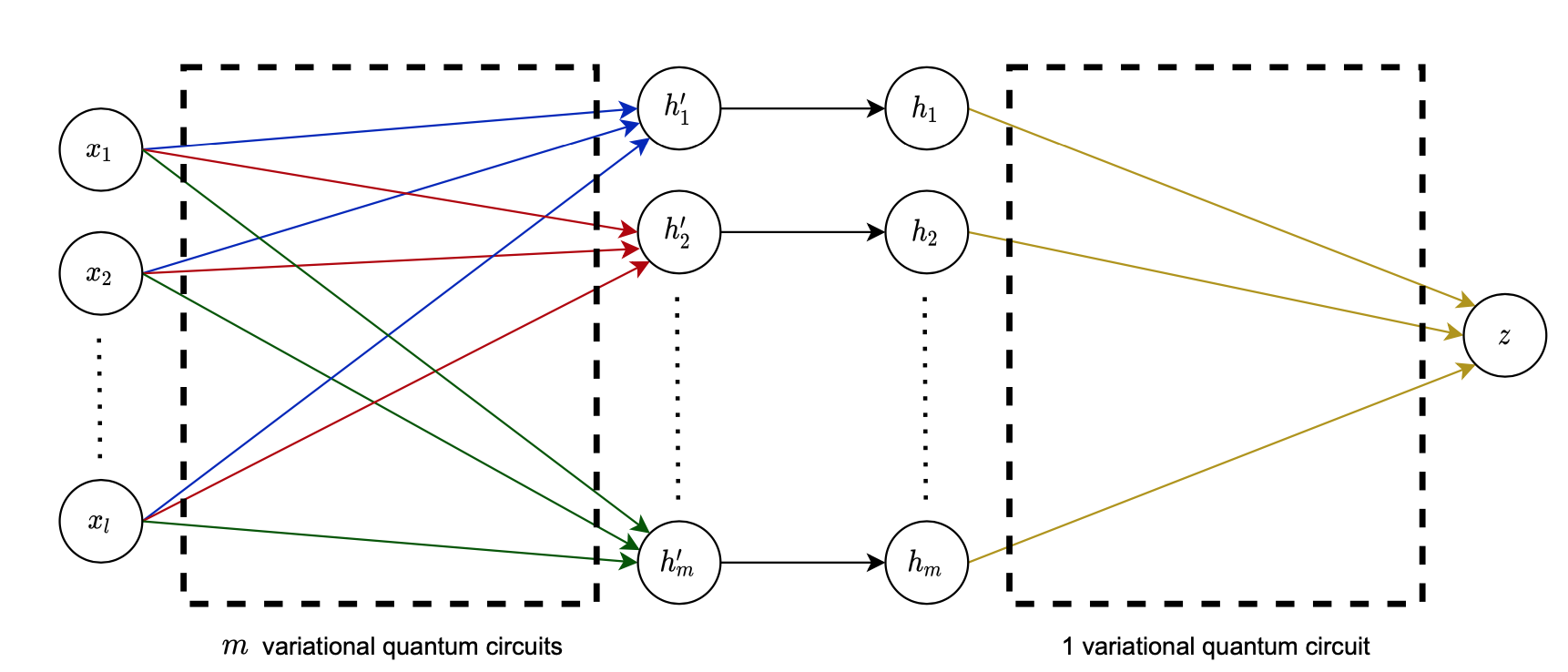}
    \caption{Architecture of a single hidden layer quantum neural network. The inputs and corresponding output of a particular variational quantum circuit are connected by arrows of the same color. The input data point $\bm{x}$ has dimension $l$. The output of the entire network is denoted by $z$.}
    \label{fig:qnn_schematic}
\end{figure}

Using the architecture shown in Figure \ref{fig:qnn_schematic}, our hybrid neural network contains $m + 1$ variational quantum circuits. The first $m$ circuits comprise the hidden layer of the feedforward network. Each of these circuits has its own parameter vector $\bm{\theta}_i^{(1)}$, and they all share the same input $\bm{x}$. The output of each circuit in the hidden-layer is stacked to create an $m$ dimensional vector $\bm{h}' \in [-1, 1]^m$. Collectively, we denote the $m$ circuits of the hidden layer as a function $f_1: [0, \pi]^d  \rightarrow [-1, 1]^m$. 
\begin{align}
    \bm{h}' = f_1(\bm{x}, \bm{\theta}^{(1)})
\end{align}

Before $\bm{h}'$ is passed to the feature map of the final variational quantum circuit, a transformation is applied so that each value is within the interval $[0, \pi]$:
\begin{align}
    \bm{h} = \frac{\pi}{2}(\bm{h}' + 1)
\end{align}
The last variational quantum circuit is run with input $\bm{h}$ and parameter vector $\bm{\theta}^{(2)}$. We denote this quantum circuit as a function $f_2:[0, \pi]^m \rightarrow [-1, 1]$. All together, the hybrid neural network is expressed by the following composite function:
\begin{align}
    f_{\text{NN}}(\bm{x}, \bm{\theta}^{(1)}, \bm{\theta}^{(2)}) = f_2 \left( \frac{1}{2}(f_1(\bm{x}, \bm{\theta}^{(1)}) + 1), \bm{\theta}^{(2)} \right)
\end{align}

The output of the hybrid neural network can be used to learn the probability distribution that a point $\bm{x}$ is labeled $y$ using the same method described by Equation \ref{eq:prob_perceptron}:
\begin{align}
    P(y|\bm{x}) = \frac{y f_{\text{NN}}(\bm{x}, \bm{\theta}^{(1)}, \bm{\theta}^{(2)}) + 1}{2}
\end{align}
Training the hybrid neural network on binary classification problems is similar to training the individual variational quantum circuit classifier. We minimize the same cost function (given by Equation \ref{eq:cost}), and we still use parameter shift rule to compute the gradient of each circuit. Now however, we must compute the gradient of the cost function with respect to the output layer's parameter vector $\bm{\theta}^{(2)}$ as well each parameter vector in the hidden layer $\bm{\theta}_i^{(1)}$. This is performed most efficiently using backpropagation.

\section{Methods}
\label{sec:methods}

 \begin{figure*}
     \centering
     \includegraphics[width=15cm]{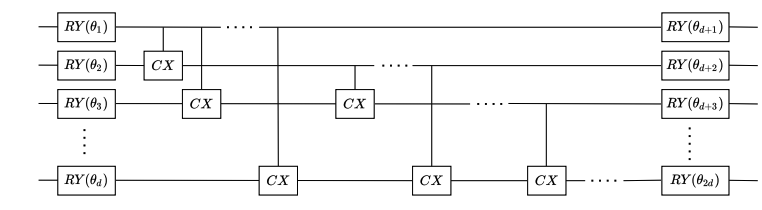}
     \caption{\textit{RealAmplitudes} ansatz from the Qiskit library with one repetition and full entanglement.}
     \label{fig:ansatz}
 \end{figure*}

\subsection{Amplitude Encoding Feature Map}
\label{sub:amplitude_encoding}
The amplitude encoding feature map is implemented by first initializing a quantum register of $d$ qubits, each in the $\ket{0}$ state. Next, a single qubit parameterized RY gate is applied to each qubit. The parameters of each gate are chosen such that the $i$th qubit is rotated by an angle $x_i$. The probability $p$ of measuring the $i$th qubit in the $\ket{0}$ state after the feature map is applied is given by the following equation:
\begin{align}
    p(x_i) = \cos^2{\left( \frac{x_i}{2} \right)}
\end{align}
Note that if $x_i$ is restricted to the interval $[0, \pi]$, $p(x_i)$ is a one-to-one function with a range spanning all possible probabilities from 0 to 1. This feature map guarantees that each unique input $\bm{x}$ will have a unique quantum encoding without requiring a large number of quantum gates.

\subsection{Real Amplitudes Ansatz}
 The \textit{RealAmplitudes} ansatz from the Qiskit library consists entirely of single qubit RY rotation gates and two qubit CX entanglement gates. First, a parameterized RY gate is applied to each qubit. The parameters of these RY gates are the first $d$ parameters of the ansatz. Next, two qubit CX gates are applied to each possible combination of qubits in the quantum state. By convention, the least significant qubit is used as the control bit each time. Finally, each qubit is subject to another parameterized RY gate. The parameters of these gates are the second $d$ parameters of the ansatz. Additional rounds of CX entanglements and RY rotations can be added to the ansatz by adjusting the number of repetitions. However, to avoid long training times, we used one repetition for each variational quantum circuit. With this specification, the \textit{RealAmplitudes} ansatz always has exactly $2d$ parameters.

\subsection{Preprocessing Real Valued Data}
As mentioned in Section \ref{sub:amplitude_encoding}, we require each feature to be within the interval $[0, \pi]$ before passing it to the feature map. For Bars and Stripes, this is not an issue since all features have a binary value. Alternatively, the two real valued data sets must be modified before training since both include many points with feature values outside of the desired range. We prepared these data sets for the variational quantum circuit using the following procedure:
\begin{enumerate}
    \item Scale the data set so that each feature has a mean of 0 and a variance of 1 using Sci-Kit Learn's \textit{StandardScalar} class.
    \item Divide the modified data set by its feature with the largest absolute value. Now all features in the data set have a value between -1 and 1.
    \item Multiply the modified data set by $\pi/2$.
    \item Add $\pi / 2$ to each feature of the modified data set.
\end{enumerate}
After this procedure is performed, each feature in the modified data set will fall within the interval [$0, \pi$], ensuring that every unique input $\bm{x}$ will have a unique quantum encoding.

\subsection{Hardware and Job Specifications}
During the simulated quantum trials, each quantum circuit was run using the IBM QASM simulator. During the actual quantum trials, each quantum circuit was run using the IBM Mumbai quantum computer or the IBM Montreal quantum computer. Both quantum computers have 27 qubits and a quantum volume of 128. They also both use CX, ID, RZ, SX, and X gates. When our results were compiled, the average CNOT error of IBM Mumbai was $8.572 \times 10^{-3}$, and the average readout error was $3.834 \times 10^{-2}$. The average CNOT error of IBM Montreal was $4.597 \times 10^{-2}$, and the average readout error was $
1.706 \times 10^{-2}$. To determine the output (expectation value) of a variational quantum circuit on a particular input, we ran the circuit 1024 times and then averaged the result of each run. Each job was initialized and sent to the quantum computer using a personal laptop with a 2.7 GHz Dual-Core Intel i5 processor and 8 GB 1,867 MHz DDR3 memory. This laptop was also used to process the results of each job and optimize model parameters accordingly.

\section{Results}
\label{sec:results}

\begin{table*}
    \centering
    \begin{tabular}{c|c|c|c|ccc|ccc|ccc|ccc}
    & & & & \multicolumn{3}{c|}{\shortstack{in sample \\ accuracy}} & \multicolumn{3}{c|}{\shortstack{in sample \\ cost}} & \multicolumn{3}{c|}{\shortstack{out of sample \\ accuracy}} & \multicolumn{3}{c}{\shortstack{out of sample \\ cost}} \\
    \hline
    hardware & data & model & parameters & median & avg. & std. & median & avg. & std. & median & avg. & std. & median & avg. & std. \\ \hline
    \multirow{2}{*}{simulated} & \multirow{2}{*}{BAS} & VQC & 8 & 100.0 & 88.89 & 12.42 & 0.55 & 0.54 & 0.04 & N/A & N/A & N/A & N/A & N/A & N/A \\ 
    & & HNN & 20 & 100.0 & 100.0 & 0.0 & 0.33 & 0.35 & 0.07 & N/A & N/A & N/A & N/A & N/A & N/A \\
    \hline
    \multirow{2}{*}{simulated} & \multirow{2}{*}{synth} & VQC & 4 & 97.5 & 85.5 & 18.34 & 0.37 & 0.46 & 0.14 & 100.0 & 86.5 & 20.13 & 0.35 & 0.43 & 0.15 \\ 
    & & HNN & 12 & 97.5 & 93.88 & 9.0 & 0.29 & 0.33 & 0.13 & 97.5 & 94.5 & 8.79 & 0.25 & 0.29 & 0.14 \\ \hline
    \multirow{2}{*}{simulated} & \multirow{2}{*}{iris} & VQC & 8 & 88.12 & 81.5 & 14.37 & 0.45 & 0.48 & 0.12 & 87.5 & 82.5 & 17.92 & 0.44 & 0.48 & 0.12 \\ 
    & & HNN & 20 & 91.25 & 89.88 & 4.24 & 0.37 & 0.39 & 0.09 & 95.0 & 91.5 & 9.23 & 0.38 & 0.39 & 0.10 \\ \hline
    \multirow{2}{*}{quantum} & \multirow{2}{*}{BAS} & VQC & 8 & 50.0 & 50.0 & 0.0 & 0.71 & 0.71 & 0.01 & N/A & N/A & N/A & N/A & N/A & N/A \\ 
    & & HNN & 20 & 25.0 & 33.33 & 11.79 & 0.71 & 0.72 & 0.11 & N/A & N/A & N/A & N/A & N/A & N/A \\ \hline
    \multirow{2}{*}{quantum} & \multirow{2}{*}{synth} & VQC & 4 & 96.25 & 82.92 & 20.65 & 0.38 & 0.46 & 0.13 & 95.0 & 90.0 & 10.8 & 0.35 & 0.4 & 0.1 \\ 
    & & HNN & 12 & 96.25 & 95.0 & 3.68 & 0.26 & 0.31 & 0.07 & 100.0 & 95.0 & 7.07 & 0.23 & 0.27 & 0.06 \\ \hline
    \multirow{2}{*}{quantum} & \multirow{2}{*}{iris} & VQC & 8 & 45.0 & 45.0 & 5.0 & 0.79 & 0.79 & 0.07 & 47.5 & 47.5 & 7.5 & 0.75 & 0.75 & 0.03 \\ 
    & & HNN & 20 & 28.12 & 28.12 & 20.62 & 0.92 & 0.92 & 0.21 & 37.5 & 37.5 & 17.5 & 0.95 & 0.95 & 0.24
    \end{tabular}
    \caption{Final binary classification results of the variational quantum circuit (VQC) and hybrid neural network (HNN) on the $2 \times 2$ Bars and Stripes data set (BAS), a synthetic two dimensional data set (synth), and a subset of the iris data set (iris). Average values are denoted by avg. and the corresponding standard deviation is denoted by std.}
    \label{tab:all}
\end{table*}

We tested the hybrid neural network on three binary classification data sets. As a point of comparison, we also trained an individual variational quantum circuit classifier on each of these data sets. We trained both models on a simulated universal quantum computer and a state of the art universal quantum computer. To achieve reasonable training times, we restricted the hybrid neural network to use only $m=2$ hidden neurons. On all three data sets, 10 simulated quantum trials and 2 to 3 actual quantum trials were performed for both quantum models. Each trial, all ansatz parameters were randomly initialized using a uniform distribution with a range of $[-\pi, \pi]$.

On modern hardware, the VQC and HNN do not offer any training time advantage over classical machine learning models. In fact, it is always possible to construct a classical multilayer perceptron that requires substantially shorter training times while achieving equal or better accuracy. Since it is unclear how quantum training time will change as quantum hardware evolves, we did not report the training times from our experiments. In general, even on the smallest data set we tested, training the VQC model can take over 30 minutes. This training time is dominated by the time required to prepare and run each quantum circuit on the quantum computer. Since the HNN model is composed of multiple VQC units, the training time of the HNN is larger than the training time of the VQC. On our test data sets, we found that the training time of the HNN was 3 to 5 times larger than the training time of the VQC.

\subsection{Bars and Stripes}

\begin{figure}[H]
    \centering
    \includegraphics[width=7.5 cm]{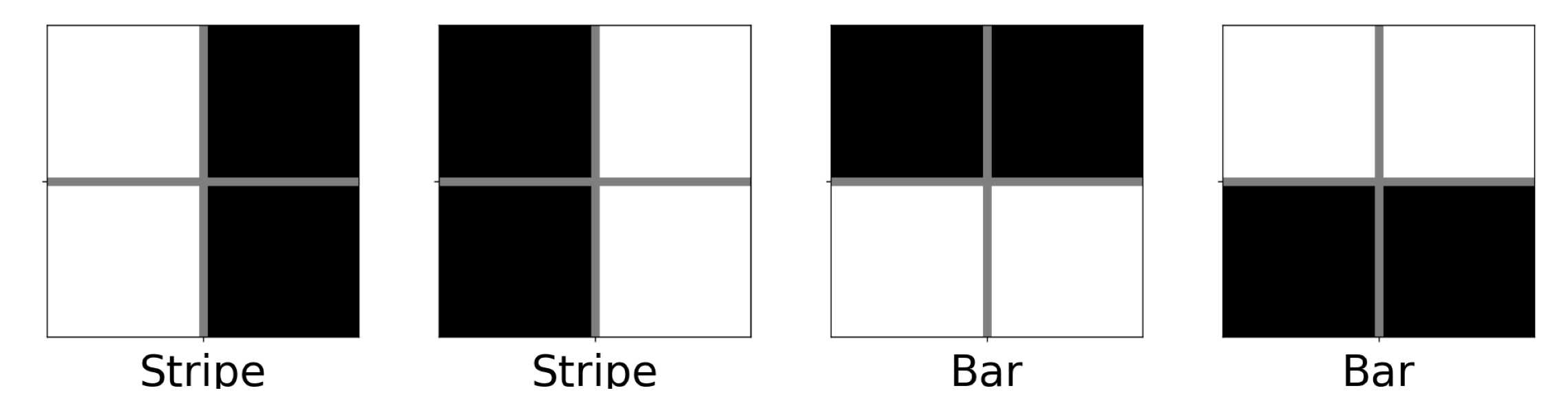}
    \caption{$2 \times 2$ Bars and Stripes data set.}
    \label{fig:bars_stripes_visual}
\end{figure}

Bars and Stripes is a synthetic data set of $n \times m$ binary black and white images. Each image in the data set is either a ``bar" or a ``stripe." A ``bar" has 1 to $m-1$ horizontal rows highlighted in black, and a ``stripe" has 1 to $n-1$ vertical columns highlighted in black. In some variations of Bars and Stripes, an entirely white image and an entirely black image is also included. We do not include these two images since their classification is ambiguous. Overall, the data set contains $N = 2^n + 2^m - 4$ images. Of these images, $N_s = 2^n - 2$ are stripes and $N_b = 2^m - 2$ are bars. An image of the $2 \times 2$ Bars and Stripes data set is depicted in Figure \ref{fig:bars_stripes_visual}.

\begin{figure}[H]
    \centering
    \includegraphics[width=7.2cm]{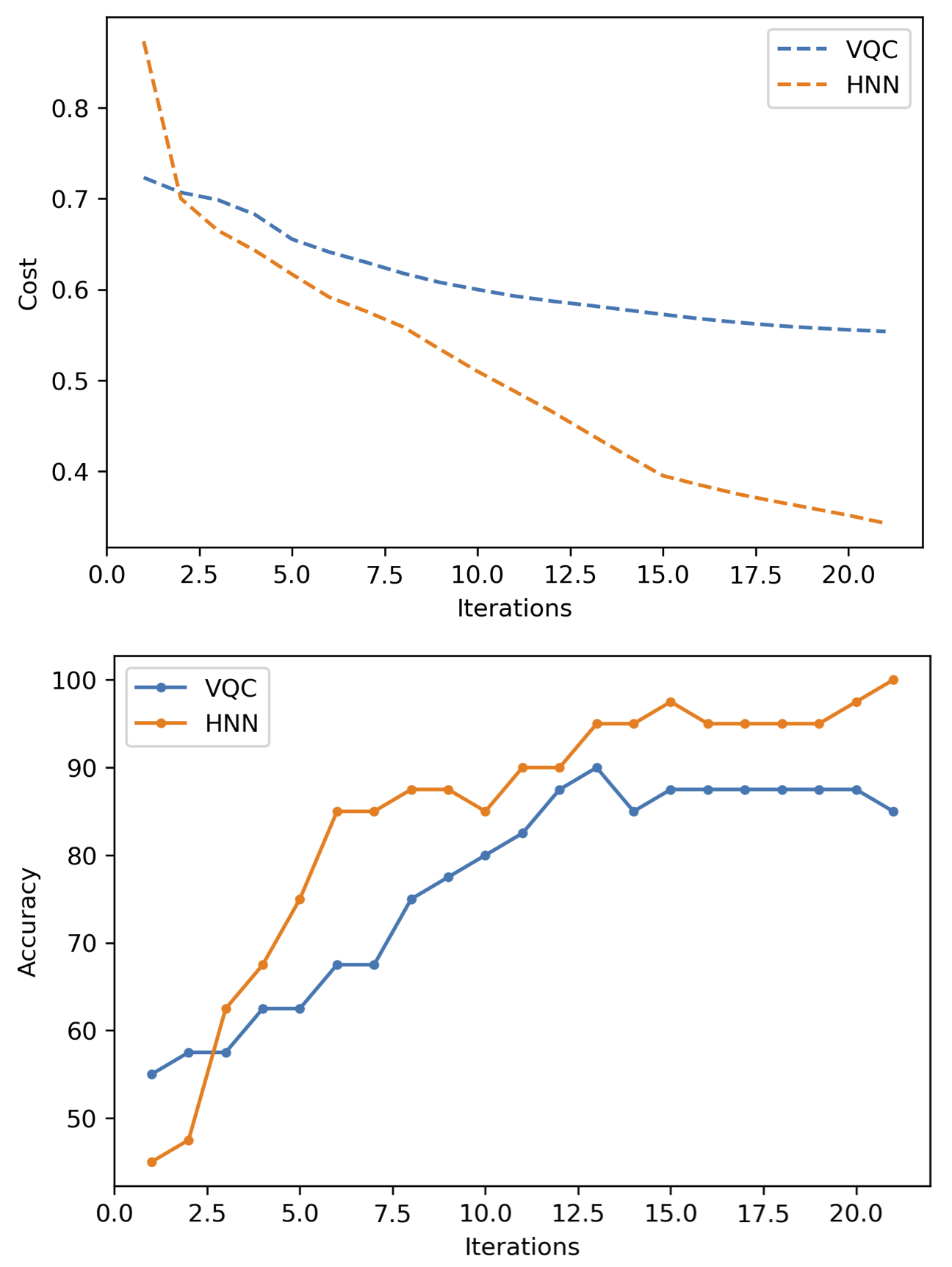}
    \caption{Average cost and accuracy achieved by the variational quantum circuit (VQC) and hybrid neural network (HNN) during training on the $2 \times 2$ Bars and Stripes data set. All illustrated trials were performed on simulated quantum hardware.}
    \label{fig:bas}
\end{figure}

We trained the variational quantum circuit classifier and hybrid quantum classical neural network on the $2 \times 2$ Bars and Stripes data set using 20 epochs of batch gradient descent with a learning rate of 0.5. All 4 points in the data set were used for training. The results of the simulated trials are reported in Table \ref{tab:all} and Figure \ref{fig:bas}. The results of the quantum trials are reported in Table \ref{tab:all}.

On simulated hardware, the HNN correctly classified every point each trial, while the VQC occasionally incorrectly classified one or more points. Nevertheless, both models achieved high accuracy on average. The slight difference in the average accuracy of each model may indicate that the HNN architecture is more resilient to unfavorable parameter initialization. On quantum hardware, both models performed poorly. The number of required qubits and gates in each model is proportional to the dimension of the data set, so it may be the case that the quantum circuits used for this data set were too large to be accurately performed on modern hardware.

\subsection{Synthetic Data}
\label{sub:synth}

\begin{figure}[H]
    \centering
    \includegraphics[width=7.5cm]{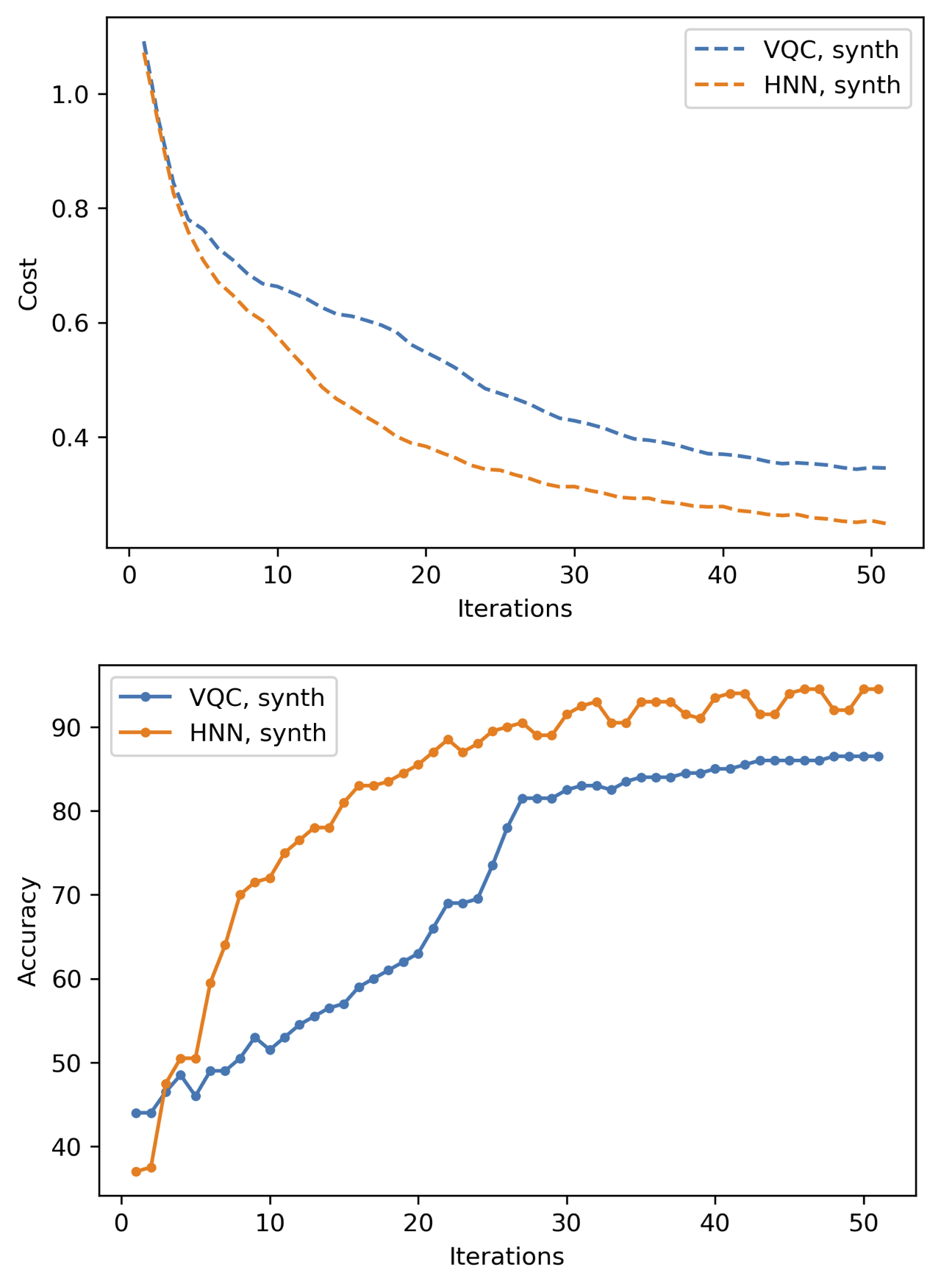}
    \caption{Average cost and accuracy achieved by the variational quantum circuit (VQC) and hybrid neural network (HNN) during training on a synthetic two dimensional data set. All illustrated trials were performed on simulated quantum hardware.}
    \label{fig:synth}
\end{figure}

We also trained the individual variational quantum circuit and hybrid quantum-classical neural network on a two dimensional, linearly separable data set generated using Sci-Kit Learn's \textit{make\_blobs()} function. This synthetic data set consisted of 100 data points split evenly between each class. In each experiment, 80 of the 100 data points were chosen at random to be used for training. Training consisted of 10 epochs of mini-batch gradient descent using a batch size of 16 points and a learning rate of 0.1. The results of the simulated trials are reported in Table \ref{tab:all} and Figure \ref{fig:synth}. The results of the quantum trials are reported in Table \ref{tab:all}.

Similar to the bars and stripes data set, the HNN and VQC both achieved high accuracy on simulated hardware. The HNN had higher average accuracy than the VQC by roughly 10 percent. Additionally, the HNN had a lower average cost than the VQC by over 30 percent. This time, both models also achieved high accuracy on actual quantum hardware. This is unsurprising since the synthetic data set is two dimensional, meaning much fewer qubits and logic gates are required in each quantum circuit.

\begin{figure}[H]
    \centering
    \includegraphics[width=7.5cm]{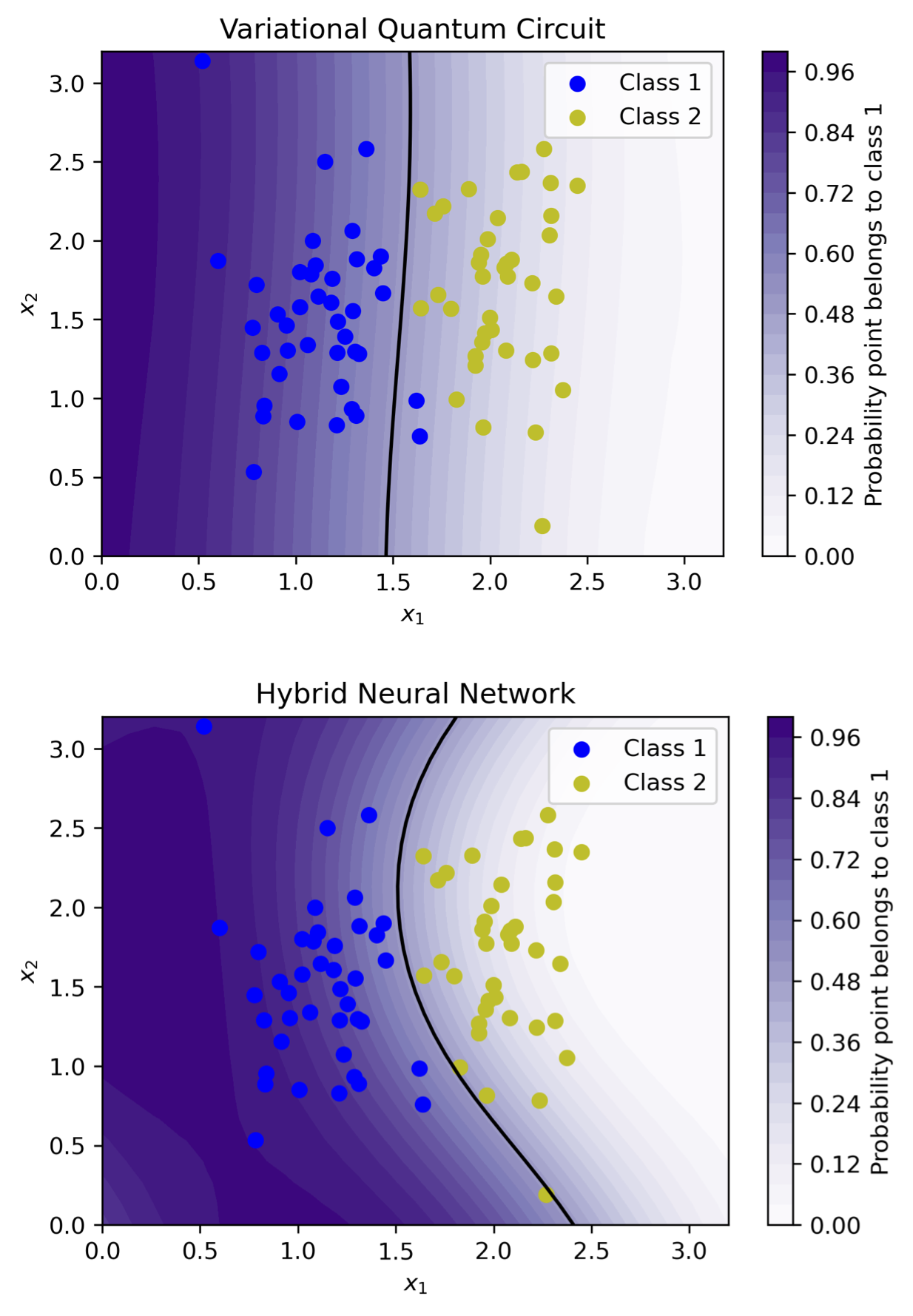}
    \caption{Final classification line and probability distribution of a variational quantum circuit and a hybrid neural network trained on a two dimensional, linearly separable data set.}
    \label{fig:illustrate_learning}
\end{figure}

Since the synthetic data set is two dimensional, it is possible to visualize the classification line and probability distribution learned by each quantum model. In Figure \ref{fig:illustrate_learning}, we have plotted this information for one of the variational quantum circuit trials and one of the hybrid neural network trials. The two examples chosen were selected because their final accuracy and final cost value were reflective of other trials of the same model type. Additionally, both examples had roughly 50\% accuracy before training.

\subsection{Iris}
\begin{figure}[H]
    \centering
    \includegraphics[width=7.5cm]{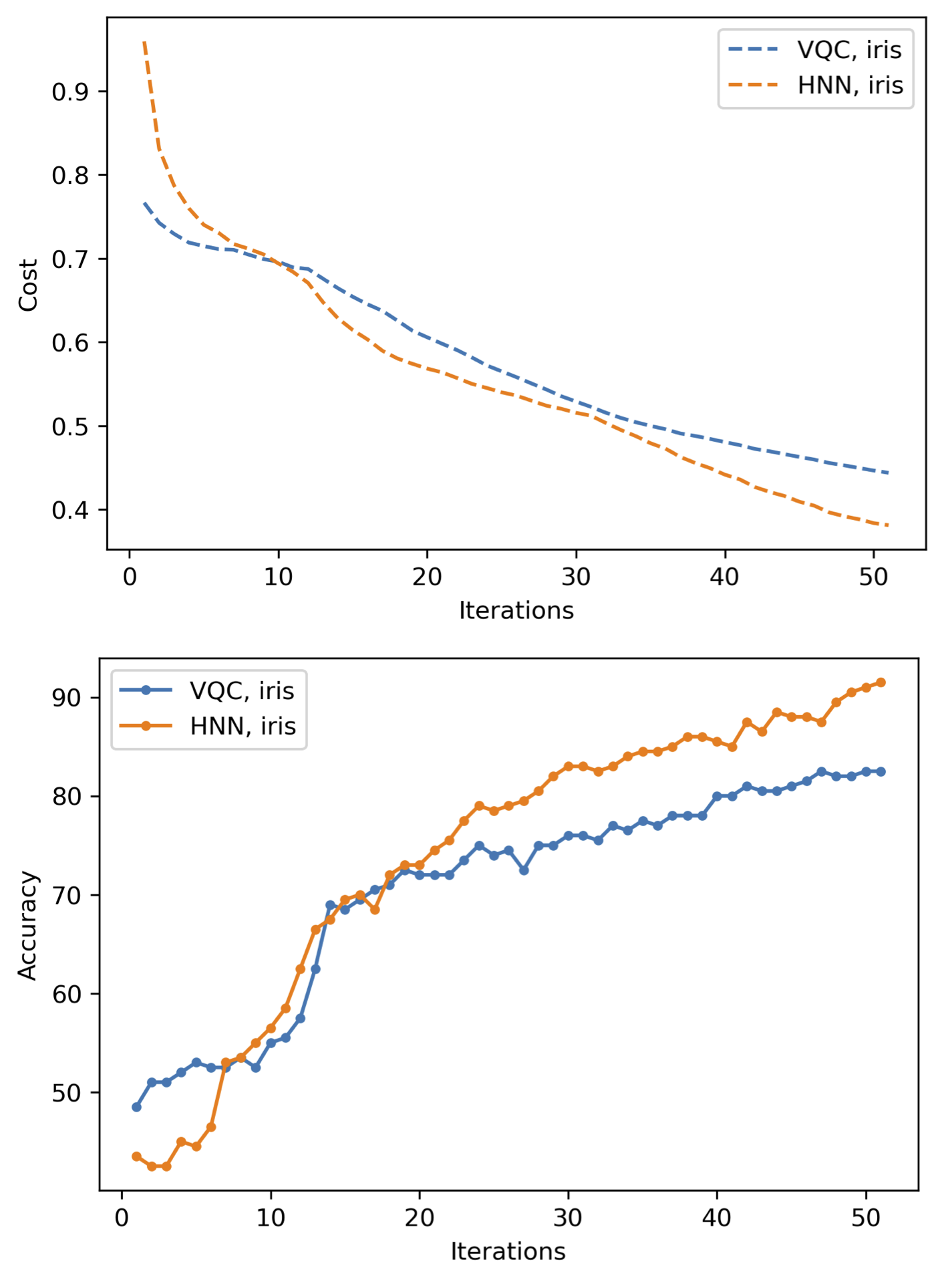}
    \caption{Average cost and accuracy achieved by the variational quantum circuit (VQC) and hybrid neural network (HNN) during training on a subset of the iris data set. All illustrated trials were performed on simulated quantum hardware.}
    \label{fig:iris}
\end{figure}

Finally, we trained the individual variational quantum circuit and hybrid quantum-classical neural network on a subset of the iris benchmark data set. The iris data set consists of 150 samples split evenly among 3 species of iris. Each iris sample is represented by four real valued features (sepal length, sepal width, pedal length, pedal width). We tested both models on the 100 samples corresponding to the iris versicolor and iris virginica species whose sample points are non-linearly separable. In each experiment, 80 of the 100 data points were chosen at random to be used for training. Training consisted of 10 epochs of mini-batch gradient descent using a batch size of 16 points and a learning rate of 0.1. The results of the simulated trials are reported in Table \ref{tab:all} and Figure \ref{fig:iris}. The results of the quantum trials are reported in Table \ref{tab:all}.

On average, the HNN achieved roughly 10 percent better accuracy than the VQC when simulated quantum hardware was used. The HNN also achieved an average cost value approximately 20\% less than the VQC. Unfortunately, like the Bars and Stripes data set, both models performed extremely poorly when quantum hardware was used. Again, we suspect the decline in performance is due to the fact the iris data set is four dimensional.

\section{Conclusion}
\label{sec:conclusion}
On simulated hardware, the hybrid quantum-classical neural network always outperformed the individual variational quantum circuit in terms of both accuracy and cost. Specifically, the average accuracy was 8 to 11 percent higher, and the average cost was 20 to 40 percent lower. Notably, the advantages achieved by the hybrid neural network were observed on both the training data set and the test data set. This suggests that they were not a product of overfitting. The learned Bernoulli distributions illustrated in Figure \ref{fig:illustrate_learning} give some indication of why the hybrid quantum-classical neural network achieves better performance. The neural network is able to produce a probability distribution with a much steeper gradient near the classification line. This enables the neural network to classify points with greater certainty than the individual variational quantum circuit, which in turn helps minimize cost. 

It is not overwhelmingly surprising that the hybrid neural network is more expressive than the variational quantum circuit classifier since it has more than twice as many parameters. Nevertheless, increasing the number of parameters of a machine learning model does not always guarantee better results, especially when data points outside of the training set are considered. At the very least, the proposed hybrid neural network architecture illustrates one effective way to add parameters to a quantum machine learning model. Some measures indicate that variational quantum circuits are more expressive than classical neural network architectures \cite{Cerezo2021}. Our proposed hybrid quantum-classical neural network architecture illustrates one approach to capitalize on these advantages when tackling more challenging machine learning tasks.

Notably, when quantum hardware was used, the variational quantum circuit classifier and the hybrid neural network both performed extremely poorly on the iris data set and the Bars and Stripes data set. This is likely because the number of qubits and number of required gates is proportional to the dimension of the data set. Increasing the number of qubits or increasing the number of gates adversely impacts the fidelity of modern quantum computation. Future research may investigate using a more sophisticated feature map or ansatz within the variational quantum circuits used by the neural network. Additionally, a more in depth study into hyper-parameter optimization of the learning rate and batch size may prove useful for improving results on modern quantum hardware. Finally, larger and more complex hybrid neural network architectures may be investigated on more challenging classification problems.

\section*{Acknowledgements}
This manuscript has been authored in part by UT-Battelle, LLC under Contract No. DE-AC05-00OR22725 with the U.S. Department of Energy. The United States Government retains and the publisher, by accepting the article for publication, acknowledges that the United States Government retains a non-exclusive, paid-up, irrevocable, world-wide license to publish or reproduce the published form of this manuscript, or allow others to do so, for United States Government purposes. The Department of Energy will provide public access to these results of federally sponsored research in accordance with the DOE Public Access Plan (http://energy.gov/downloads/doe-public-access-plan). This research used resources of the Oak Ridge Leadership Computing Facility, which is a DOE Office of Science User Facility supported under Contract DE-AC05-00OR22725.
This work was funded in part by the DOE Office of Science, High-energy Physics Quantised program.
This work was funded in part by the DOE Office of Science, Advanced Scientific Computing Research (ASCR) program.

\nocite{*}

\bibliography{bibliography}

\end{document}